 \documentclass[sigconf]{acmart}

\usepackage{booktabs} 
\usepackage{graphicx}
\usepackage{subcaption}
\usepackage{amsmath}
\usepackage{listings}
\usepackage{hyperref}

\usepackage{algorithm}
\usepackage{algpseudocode}
\usepackage{pifont}

\usepackage{multirow}
\usepackage{capt-of}
\usepackage{varwidth}
\usepackage[switch]{lineno} 
\usepackage{pythonhighlight}
\lstnewenvironment{PythonNoBorder}[1][]{\lstset{style=mypython, basicstyle=\small, frame=none, #1}}{}
\lstnewenvironment{PythonBorder}[1][]{\lstset{style=mypython, basicstyle=\small, #1}}{}

\DeclareMathAlphabet\mathbfcal{OMS}{cmsy}{b}{n}

\AtBeginDocument{%
  \providecommand\BibTeX{{%
    \normalfont B\kern-0.5em{\scshape i\kern-0.25em b}\kern-0.8em\TeX}}}

\begin{document}
\title[BOLT: An Efficient Deep Learning Framework for CPUs]{BOLT: An Automated Deep Learning Framework for Training and Deploying Large-Scale Search and Recommendation Models on Commodity CPU Hardware}



\author{Nicholas Meisburger}
\author{Vihan Lakshman}
\author{Benito Geordie}
\author{Joshua Engels}
\email{nicholas@thirdai.com}
\affiliation{%
  \institution{ThirdAI}
  \city{Houston}
  \state{Texas}
  \country{USA}
}

\author{David Torres Ramos}
\author{Pratik Pranav}
\author{Benjamin Coleman}
\author{Benjamin Meisburger}
\affiliation{%
  \institution{ThirdAI}
  \city{Houston}
  \state{Texas}
  \country{USA}
}

\author{Shubh Gupta}
\author{Yashwanth Adunukota}
\author{Siddharth Jain}
\author{Tharun Medini}
\author{Anshumali Shrivastava}
\affiliation{%
  \institution{ThirdAI}
  \city{Houston}
  \state{Texas}
  \country{USA}
}


\copyrightyear{2023}
\acmYear{2023}
\setcopyright{acmlicensed}\acmConference[CIKM '23]{Proceedings of the 32nd ACM International Conference on Information and Knowledge Management}{October 21--25, 2023}{Birmingham, United Kingdom}
\acmBooktitle{Proceedings of the 32nd ACM International Conference on Information and Knowledge Management (CIKM '23), October 21--25, 2023, Birmingham, United Kingdom}
\acmPrice{15.00}
\acmDOI{10.1145/3583780.3615458}
\acmISBN{979-8-4007-0124-5/23/10}

\renewcommand{\shortauthors}{Nicholas Meisburger et al.}

\begin{abstract}
Efficient large-scale neural network training and inference on commodity CPU hardware is of immense practical significance in democratizing deep learning (DL) capabilities. Presently, the process of training massive models consisting of hundreds of millions to billions of parameters requires the extensive use of specialized hardware accelerators, such as GPUs, which are only accessible to a limited number of institutions with considerable financial resources. Moreover, there is often an alarming carbon footprint associated with training and deploying these models. In this paper, we take a step towards addressing these challenges by introducing BOLT, a sparse deep learning library for training large-scale search and recommendation models on standard CPU hardware. BOLT provides a flexible, high-level API for constructing models that will be familiar to users of existing popular DL frameworks. By automatically tuning specialized hyperparameters, BOLT also abstracts away the algorithmic details of sparse network training. We evaluate BOLT on a number of information retrieval tasks including product recommendations, text classification, graph neural networks, and personalization. We find that our proposed system achieves competitive performance with state-of-the-art techniques at a fraction of the cost and energy consumption and an order-of-magnitude faster inference time. BOLT has also been successfully deployed by multiple businesses to address critical problems, and we highlight one customer case study in the field of e-commerce. 
\end{abstract}



\maketitle

\section{Introduction}

In recent years, extremely large-scale neural networks have dramatically altered search and recommendation systems. However, this shift towards building ever-larger models has raised a number of challenges associated with training and deployment. First and foremost, these massive search and recommendation models are characterized by their \textbf{high-dimensional output spaces}, and often contain hundreds of millions if not billions of parameters. Training such networks using standard deep learning frameworks requires the use of costly specialized hardware such as GPUs and TPUs, exacerbating the gulf between institutions with the resources to build these models and those without such capabilities. Additionally, these models need to be retrained frequently as new data is generated from user interactions, thus necessitating additional resources and cost.

Secondly, \textbf{low inference latency} is crucial in search and recommendation settings as these models are deployed to serve real-time user queries or interactions. This requirement makes deploying such models a difficult engineering challenge; practitioners often must shrink the network using compression techniques such as knowledge distillation~\cite{hinton2015distilling}, quantization~\cite{han2015deep}, and pruning~\cite{blalock2020state}, which can also significantly degrade the model's quality. Finally, there is an alarming \textbf{energy cost and carbon footprint} associated with training and deploying these models~\cite{strubell2019energy}, driven in large part by the fact that the majority of deep learning inference cycles at large web companies are devoted to search and recommendations \cite{gupta2020architectural}.

Motivated by these considerable computational barriers to realizing the full promise of large-scale neural networks for retrieval tasks, we introduce BOLT, a modular deep learning framework for large-scale search and recommendation problems that can train models with billions of parameters on cost-effective CPU hardware with inference latencies on the order of a few milliseconds. BOLT achieves these breakthroughs through \emph{algorithmic} advances on the classical neural network training approach. In particular, we provide a commercial-grade implementation of the SLIDE algorithm of~\cite{spring2017scalable,chen2020slide}, which uses adaptive sparsity to avoid expensive dense matrix multiplication. BOLT also provides a number of features that, to our knowledge, are novel amongst deep learning frameworks, including the ability to configure network sparsity to gracefully trade training time for model quality, automated tuning of specialized hyperparameters, and fast sparse inference in deployment. \textsc{BOLT} has also been tested and deployed by several companies in production environments, validating the reliability and performance of the system.

In summary, we make the following contributions in this paper:
\begin{enumerate}
    \item \textbf{Production-Grade System:} We present BOLT, a production-grade library for building neural networks specialized for high dimensional output spaces that leverages sparse computations to efficiently train and predict with large-scale models on standard CPU hardware as opposed to costly hardware accelerators. To our knowledge, \textsc{BOLT} is the first production-grade library to implement recent advances in sparse training via locality sensitive hashing primitives.
    \item \textbf{Sparsity-First Development Features:} We introduce a variety of novel features for sparse neural network training that further enhance the functionality and utility of \textsc{BOLT}. Specifically, BOLT features autotuning for sparsity related hyperparameters to simplify integrating sparsity in model training and deployment. We also develop a novel \emph{sparse inference} setting which further accelerates prediction speed at a negligible cost to model quality.
    
    \item \textbf{Real-world Impact:} Our work provides the first comprehensive evaluation of LSH-based sparse neural networks on a variety of tasks and high-performing baselines in the literature. Multiple organizations have also successfully trained and deployed BOLT models for critical business applications. In this paper, we provide a case study summarizing our journey deploying BOLT into production to improve search relevance at Wayfair, a leading online furniture retailer. 
    
\end{enumerate} 



\section{Related Work}

\subsection{Sparse Neural Network Training}

Sparsity plays a central role in the scientific study of deep learning. A number of works have explored efficiently pruning neural networks to achieve improvements in memory footprint and inference speed~\cite{sanh2020movement, frankle2020pruning, price2021dense, molchanov2016pruning, zhao2019variational}. However, to our knowledge, all of these prior pruning approaches still involve performing the model training on specialized hardware such as GPUs, leaving the efficiency gains only for the inference phase. By contrast, we show that BOLT is capable of training large networks from scratch on CPUs directly by leveraging similar principles of sparsity.



\subsection{Hashing for Deep Learning}
In conjunction with recent developments in neural network pruning and sparsity, hashing and randomization have demonstrated tremendous promise in scaling deep learning. For example, \cite{spring2017scalable} proposes to use hashing to select the neurons with the largest activations during training in order to reduce the amount of computation. Along similar lines, \cite{chen2020mongoose} leverages locality sensitive hashing to efficiently update network parameters during training. Furthermore, the aforementioned SLIDE algorithm provides a culmination of many of these early approaches by demonstrating the feasibility of training large feedforward neural networks directly on CPU hardware. We discuss the details of SLIDE further in the next section.

\section{Background: The SLIDE Algorithm}
\label{section:slide}

The SLIDE algorithm stems from the observation that, while GPUs are memory bound, CPUs are limited by throughput rather than memory, particularly for deep learning workloads. The sheer number of repeated operations involved in training deep learning models ensures the process is well suited to GPUs. Being able to efficiently reduce the computational requirements of deep learning, possibly at the expense of additional memory overhead, would make CPUs more competitive for deep learning. Since CPUs are much less expensive than GPUs, such a capability would make deep learning widely affordable and available.

\noindent\textbf{Sparsifying a Fully Connected Layer:}
The authors of SLIDE~\cite{chen2020slide} propose an algorithmic change to a standard fully connected layer to achieve this. SLIDE dynamically samples neurons most likely to have a high activation for each input. By only computing activations for these neurons it provides an accurate estimate of the activation pattern of the layer while significantly reducing computations in the forward and backward pass. This in turn reduces the total computational cost of training the model. SLIDE uses a similarity search index to perform sampling: given an input activation pattern, SLIDE queries the index to find neurons with a similar weight pattern, which are precisely the neurons likely to have a high activation. Because the sampling is done in an input-dependent way, we refer to it as dynamic sparsity. Dynamic sparsity improves upon static techniques such as pruning because it preserves the expressive power of wide layers and allows the model to train more parameters with a low computational cost.

\noindent\textbf{Locality Sensitive Hashing:}
The SLIDE algorithm uses a technique called Locality Sensitive Hashing (LSH) to build the similarity search index. LSH is a very well-studied technique that was originally introduced to break the curse of dimensionality in near-neighbor search~\cite{indyk1998approximate,andoni2008near,andoni2015optimal,andoni2014beyond,10.1145/997817.997857}. LSH has recently emerged as a sampling tool for efficient unbiased statistical estimation~\cite{spring2017scalable,charikar2017hashing,backurs2019space,siminelakis2019rehashing}. In an LSH function the probability that two points in the input space collide is proportional to the similarity between the points. The central insight of LSH-based algorithms is that given a set of points $S$, we can precompute the hashes of each point in $S$. Then, given an input $x$, we can efficiently sample similar elements $y \in S$ by hashing $x$ with the same LSH function and searching within the colliding points. 



\begin{figure}
    \centering
    \includegraphics[scale=0.42]{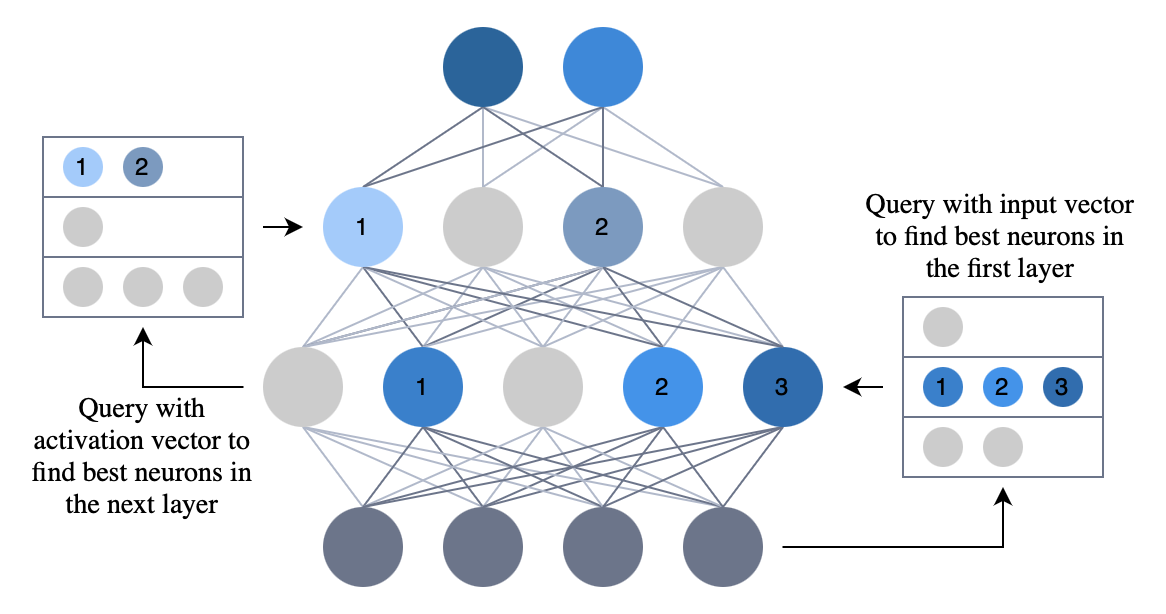}
    \caption{Overview of the SLIDE algorithm}
    \label{fig:slide}
    \vspace{-0.5cm}
\end{figure}

\noindent\textbf{Why Similarity Search Works:}
The SLIDE algorithm indexes each weight vector $w_i$ into an LSH table. To compute the output of a neural network layer, the inputs are hashed using the same hash function to retrieve a (very small) set of neurons, which are used to compute the output of the layer (see diagram \ref{fig:slide}).

The returned neurons are likely to have a high activation by the following argument. Recall that neuron activations are computed as 
$$a_i = f(\mathbf{w_i} \cdot \mathbf{x} + b_i)$$
where $f$ is the activation function, $\mathbf{w_i}$ is the weight vector of the neuron, $\mathbf{x}$ is the input, and $b_i$ is the bias of the neuron. If we assume $f$ is non-decreasing (a property of all common activation functions), then $a_i$ is non-decreasing with respect to $w_i \cdot \mathbf{x}$. We can expand this dot product as $$\mathbf{w_i} \cdot \mathbf{x} = \| \mathbf{w_i} \| \| \mathbf{x} \| \cos(\theta)$$ where $\theta$ is the angular distance between $\mathbf{w_i}$ and $\mathbf{x}$. Thus, under mild uniformity assumptions on the magnitudes of the weight vectors $w_i$, the dot product (and activation) will be largest when $cos(\theta)$ is largest, which occurs for neurons whose weight vector is closest in angular distance to the given input vector.

This fact allows us to use well-established LSH techniques that are sensitive to angular similarity~\cite{goemans1994879,charikar2002similarity,li2012one,shrivastava2014improved} to identify elements with large activations. 


\section{BOLT}

BOLT is designed to enable production scale search and recommendation workloads on CPUs through algorithmic acceleration. It achieves this by using recent methods such as SLIDE ~\cite{chen2020slide}, as well as other efficient algorithms that leverage sparsity, as fundamental computational primitives. BOLT provides a simple interface and abstracts away the additional hyperparameters associated with these techniques. It allows users to define tasks and models with a simple high level API, and then internally uses automated hyperparameter tuning to select the appropriate algorithm, sparsity, etc. to maximize the performance of the model. 


\subsection{Core Library}

In addition to common operations (Ops) associated with a modern deep learning framework BOLT also implements Ops which leverage sparsity and other algorithmic optimizations to speedup computations. Based on the model and task defined by the user, BOLT internally creates an efficient computation graph composed of Ops, with optimized sparse Ops used when possible to speedup training and inference. This model is then trained using BOLT's autograd capability which can make additional optimizations due to sparsity. During backpropagation BOLT's autograd can traverse only used links between neurons in a model, which can greatly optimize performance with sparse Ops. Additionally BOLT can track memory accesses to parameters in sparse Ops such that when applying parameter updates it only needs to update parameters that were used for the given training batch.

\subsection{Automated Sparsity Hyperparameter Tuning}

BOLT automatically tunes sparsity related hyperparameters to achieve high performance. As an example, we present a novel method to tune the hyperparameters of a sparse fully connected layer. This method forms the core of one of our optimizations to the SLIDE layer described in Section~\ref{section:slide}. 

At a high level, the intuition for our analysis is as follows: given an input activation pattern, each LSH table identifies a set of neurons (see Figure~\ref{fig:slide}). In order to have enough neurons to meet the user-specified sparsity level, we need to ensure that we match enough neurons. We can do this by having multiple LSH tables, but this adds overhead, so we want to find the minimum number of tables needed to get enough neurons. We can do this by making some load balancing assumptions about the LSH table. 

Consider a sparse layer with dimension $d$ and sparsity $s$, where sparsity is defined as the ratio of neurons whose activation we explicitly evaluate. We will hash each of the $d$ weight vectors into $L$ LSH tables, and each table will have range $2^K$. Let $X_{i, j}$ be the random variable that represents the number of neurons in bucket $i$ of table $j$. If we assume that the weight vectors are well distributed, then each bucket in each table should have $d / 2^K$ elements on expectation, or in other words for all $i, j$,
$$E(X_{i, j}) = \frac{d}{2^K}$$
Given an input activation pattern, we will collect all of the neurons from one bucket in each table. Ideally, we want the union of these buckets to contain at least $sd$ neurons. Since there are $L$ tables, and the expected number of neurons in each bucket is $\frac{d}{2^K}$, in order to have enough matched points to return we want
$$\frac{Ld}{2^K} \ge sd$$
We now introduce a "safety factor" scalar $c_1$, which represents how many times larger the right side of the inequality (the expected number of neurons returned) is greater than $sd$ (the desired number of neurons returned). If $c_1 \ll 1$, we may run out of neurons, while if $c_1 \gg 1$, we will have many unnecessary LSH tables. Thus, our final equation is
$$L = c_1s2^K$$
Both $L$ and $K$ are free variables here, so we now introduce an additional equation based on the cost of maintaining the hash tables: let $d_{\mathrm{prev}}$ be the dimension of the previous layer and $d$ be the dimension of the current sparse layer. Then the cost of hashing is $KLd_{\mathrm{prev}}$ and the cost of evaluating the chosen neurons is $sdd_{\mathrm{prev}}$, while the cost of evaluating all neurons (a dense computation) is $dd_{\mathrm{prev}}$. Thus, choosing a minimum speedup ratio $c_2$ where $c_2 < 1$, we require 
$$KLd_{\mathrm{prev}} + sdd_{\mathrm{prev}} \le c_2 d_{\mathrm{prev}}d \implies KL + sd \le c_2d$$ 
We find that larger $L$ up to about $L = 256$ gives better results (since looking at more hash tables averages the randomness from each table and increases the quality of each table of a higher quality), so our final goal is to maximize $L$ subject to 
$$KL + sd \le c_2d \qquad L \le 256 \qquad L = c_1s2^K $$
To set $K$ and $L$ in practice, we choose $c_1 = 1$, which in our experiments gives enough neurons greater than $95$\% of the time, and $c_2 = 0.1$, which corresponds to a potential $10$ times speedup. We then try substituting increasing integer values of $K$ into the third equation to solve for $L$, continuing this process until the first equation is no longer satisfied. To help load balance the hash tables we also cap the maximum number elements in each hash bucket at $R$. We find that setting $R$ equal to twice the expected number of neurons in each bucket sufficiently load balances the hash tables without a noticeable impact on performance (weight vectors are frequently well distributed).

To validate this strategy we conducted a grid search on the Amazon-670K dataset ~\cite{Bhatia16} for the first 3 epochs of training with different values of $K$, $L$, and $R$. We found that our autotuning achieved an accuracy of 1 absolute percent error against the best parameters found by the grid search, whereas many of the other combinations of parameters yielded results that had 3-5 absolute percent error compared against the best configuration.




\subsection{Sparse Inference}

Dynamic sparse deep learning methods typically use sparsity to speed up training but disable it during inference~\cite{chen2020slide, chen2020mongoose}. However, recent work has explored using similarity search indices to dynamically choose neurons during inference~\cite{liu2020climbing}. BOLT builds on this research and supports dynamic sparsity in inference to reduce latency.

Recall that during training, we can use LSH tables to quickly find high activation neurons corresponding to a training sample's activation pattern. The authors of ~\cite{liu2020climbing} examine using these hash tables for inference in the same way as training and find that the correct neurons are not always returned. To minimize the probability of this event occurring, we introduce two strategies. The first strategy is to increase the inference sparsity while keeping LSH tables the same; this method evaluates more neurons, increasing the chance that the neuron corresponding to the correct class is returned at the cost of increasing the computational cost. The second strategy is during training, when the hash buckets containing a correct label are not selected, we insert that label into the hash buckets that were selected instead. This increases the chance of the correct label being retrieved for similar samples in the future.

We report the results of using sparse inference and dense inference on Amazon 670k in Table~\ref{table:sparse_inference} after training for $5$ epochs. We keep the inference sparsity the same as the training sparsity ($0.05$ for the output layer), and report results with both adding label neurons (ALN) and not adding label neurons to buckets during training. We include TensorFlow and PyTorch as well for a comparison. Interestingly, we find that \textit{both} sparse and dense inference do better when we add label neurons to buckets during training; we hypothesize that adding label neurons to buckets reduces the number of non-label neurons we select during sparse training, and thus speeds up convergence.

\begin{table}[h]
\begin{tabular}{ |c|c|c|}
 \hline
 Dataset & Accuracy & Inference Time (ms) \\ \hline
 BOLT Sparse Inference (ALN) & 0.345 & 4.4\\ \hline
 BOLT Dense Inference (ALN) & 0.348 & 63\\ \hline
 BOLT Sparse Inference & 0.298 & 4.0\\ \hline
 BOLT Dense Inference  & 0.325 & 67\\ \hline
 Tensorflow-CPU  & 0.346 & 44.4\\ \hline
 PyTorch-CPU & 0.341 & 27.4\\ \hline
 Tensorflow-GPU & 0.346 & 1.9\\ \hline
 PyTorch-GPU & 0.349 & 0.6 \\ \hline
\end{tabular}
\caption{\label{table:sparse_inference} Inference on Amazon 670k after $5$ epochs of training.}
\vspace{-0.8cm}
\end{table}

\begin{figure*}[t]
    \includegraphics[scale=0.65]{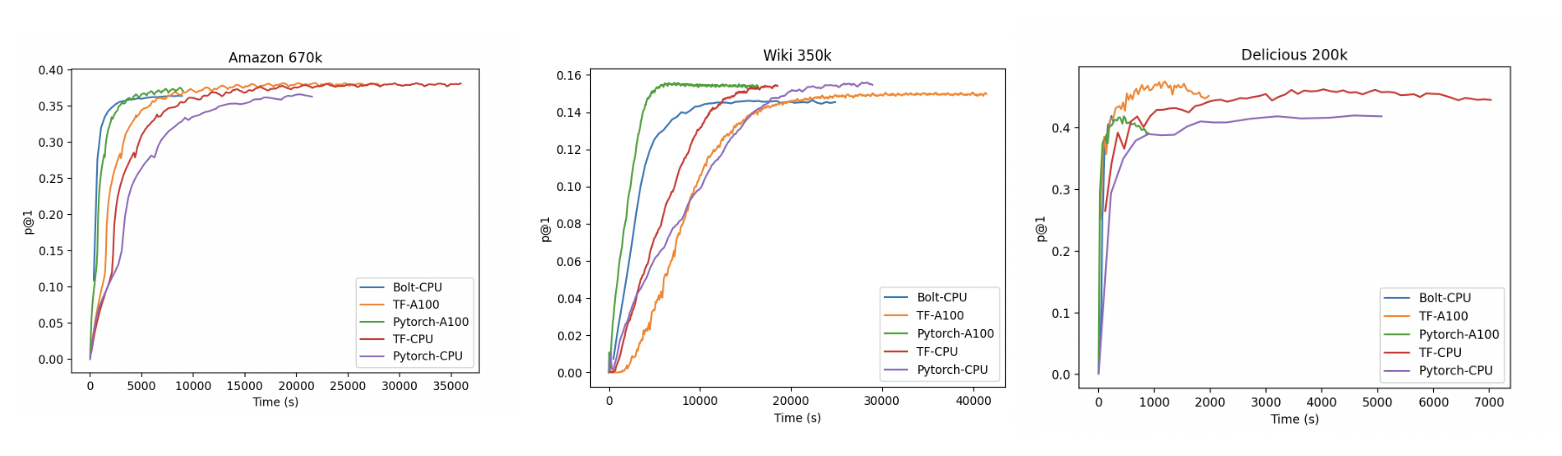}
    \vspace{-0.7cm}
    \caption{Precision@1 vs Time for the Amazon-670K, Wiki-350K, and Delicious 200K extreme classification recommendation benchmarks. Up and to the left is better. We observe that BOLT on a CPU tends to perform on par with TensorFlow and PyTorch models trained on a much more powerful A100 GPU. We also see that the advantages of BOLT emerge as we increase the number of output classes (right to left), which validates BOLT's strategy of sparsely computing activations.}
    \label{fig:extreme_classification}
\end{figure*}

\begin{figure*}[t]
    \centering
    \includegraphics[scale=0.377]{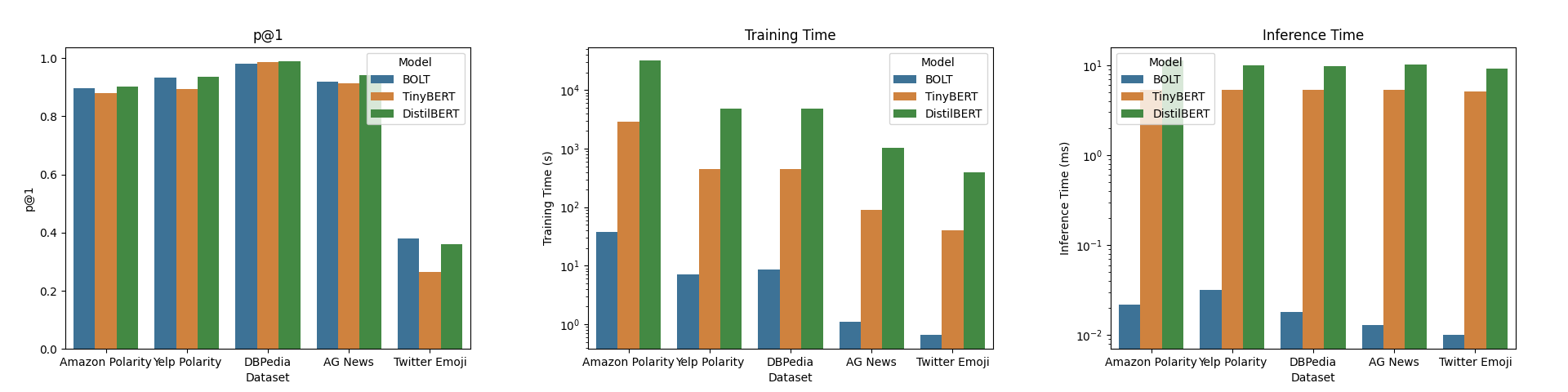}
    \vspace{-0.7cm}
    \caption{Low-Latency Text Classification Results. We observe that BOLT achieves comparable accuracy to distilled transformer models while achieving reduced training times and lower inference latencies. The TinyBERT and DistilBERT timing numbers are measured on an A100 GPU while BOLT utilizes a c6i.8xlarge CPU instance.}
    \label{fig:text_classification}
\end{figure*}

\section{Case Studies}

We will now examine case studies that showcase BOLT's performance in a variety of search and recommendation settings. We instantiate BOLT using variants of the SLIDE model architecture framework discussed earlier. We do not disclose our precise architectures and associated parameters to preserve business confidentiality, but do provide code to reproduce these results via our high-level BOLT Python API \footnote{\url{https://github.com/ThirdAIResearch/BOLT_Benchmarks}}. Unless otherwise stated, we train and evaluate all CPU-based experiments on an AWS c6i.8xlarge instance with 16 physical cores and 64 GB RAM and all GPU experiments on a Paperspace NVIDIA A100 machine with 80GB of GPU memory. Our inference latency results were all obtained by taking the average inference latency on 1000 data points.

\subsection{Extreme Classification}

Extreme classification is a machine learning problem where the output label space is considerably large (typically greater than $100,000$). This setting frequently appears in search and recommendation contexts and typical problem domains include product search and document search. For these experiments, we trained the same model using BOLT, Tensorflow GPU/CPU, and PyTorch CPU/GPU on three datasets from the Extreme Classification Repository~\cite{Bhatia16}. The results of these experiments in Figure~\ref{fig:extreme_classification} show BOLT achieves comparable performance to Tensorflow and PyTorch on an A100 GPU and is considerably faster than either engine on CPU. We note that the NVIDIA A100 processor is a particularly strong baseline for this case study since, unlike earlier generations of GPUs, it includes specific support for sparsity \cite{nvidia-sparsity}. Nevertheless, we find that BOLT achieves comparable performance on a CPU machine available at a 3-10x fraction of the cost. 


\subsection{Text Classification in the Low-Latency Regime}

In this section, we conduct experiments comparing BOLT's performance on text classification tasks against popular transformer models \cite{vaswani2017attention} that are optimized for low latency and faster fine tuning. Specifically, we compare against TinyBERT~\cite{turc2019well} and DistilBERT~\cite{sanh2019distilbert} using their PyTorch implementations in the HuggingFace transformers library \cite{wolf2019huggingface}. We use the pre-trained versions of the transformer models for initialization, while we train BOLT from scratch. Moreover, we measure the training time and inference speed of the transformer baselines on an A100 GPU while we use a c6i.8xlarge AWS CPU for BOLT. The training time numbers presented for the transformer models are only the fine tuning time; we do not include any pre-training time. For all models, the training time represents the time required for the model to train or fine tune with 3 passes over the data, and the p@1 results are the best test accuracy achieved by the model during these 3 epochs of training. We evaluated these models on the Amazon Polarity ~\cite{zhangCharacterlevelConvolutionalNetworks2015, mcauley2013hidden}, Yelp Polarity ~\cite{zhangCharacterlevelConvolutionalNetworks2015}, DBPedia ~\cite{zhangCharacterlevelConvolutionalNetworks2015, lehmann2015dbpedia}, AG News ~\cite{zhangCharacterlevelConvolutionalNetworks2015}, and Twitter Emoji ~\cite{barbieri2020tweeteval} datasets. We accessed each of these benchmarks through the HuggingFace datasets library~\cite{lhoest2021datasets}. The p@1, training time, and inference time are summarized in figure \ref{fig:text_classification}.

\subsection{Personalized Recommendations}


Recent research in personalization has obtained state-of-the-art results by treating user history as an ordered sequence (as opposed to an unordered set) and by leveraging other features such as user metadata, item metadata, or images~\cite{sasrec, sun2019bert4rec, Rashed_2022}. Since traditional methods like matrix factorization cannot handle sequential information, practitioners have turned to deep learning techniques. However, state-of-the-art deep learning methods rely on expensive transformer models, rendering them infeasible for training and inference on CPUs. We implemented a personalized recommendation engine with less compute by transforming sequential features into a high dimensional sparse vector format tailored to BOLT’s strengths.

We evaluated our personalized recommendation system on the next item prediction problem: given a user's interaction history, predict the next item they will interact with out of all items. The test set consists of the last interaction of every user while all prior actions are used for training. Movielens1M~\cite{harper2015movielens}, Amazon Games~\cite{chin2018anr}, and Netflix100M~\cite{cremonesi2010performance} are popular datasets for this problem as they consist of chronologically sorted records of interactions in the format:

\begin{verbatim}
user_id,item_id,timestamp,other_features,...
\end{verbatim}

Since we are interested in methods that are suitable on a CPU, we used a two-tower TensorFlow Recommender (TFRecO model) as a baseline instead of transformer-based sequential models. TFRec has a tendency to rank seen items higher, but repeated interactions are rare, so we augmented TFRec's output by removing seen items from the recommended list. In the following two tables, we present the recall and end to end inference latency of the two models. Latency includes the filtering step for TFRec (each user's "seen set" is precomputed) and includes the data preprocessing step for BOLT.


\begin{table}[h]
\begin{tabular}{|c|c|c|}
 \hline
 Dataset & BOLT & TF Recommender \\ \hline
 Movielens1M & 0.240 & 0.03642\\ \hline
 Amazon Games & 0.134 & 0.0193\\ \hline
 Netflix100M & 0.0661 & 0.00759\\ \hline
\end{tabular}
\captionof{table}{Personalized recommendation recall@10}\label{personalization-acc}
\vspace{-1.1cm}
\end{table}
\begin{table}[h]
\begin{tabular}{|c|c|c|}
 \hline
 Dataset & BOLT & TF Recommender \\ \hline
 Movielens1M & 1ms & 56ms\\ \hline
 Amazon Games & 10ms & 117ms\\ \hline
 Netflix100M & 7ms & 92ms\\ \hline
\end{tabular}
\captionof{table}{Personalized recommendation inference latencies}\label{personalization-inf}
\vspace{-0.5cm}
\end{table}


\subsection{Graph Learning}
In recent years, graph neural networks have become widely utilized in recommender systems \cite{wu2022graph}. In this case study, we focus on a specific type of graph learning problem: node classification. Following recent work that examined Non-Homphilous graphs (graphs where neighbors are not necessarily likely to be the same class) \cite{lim2021large} \cite{lim2021new}, we integrated BOLT into the Non-Homophilous Graph Benchmarks suite \cite{lim2021large}. In Table~\ref{graph-learning}, we compared BOLT against a subset of methods on the YelpChi, Pokec, and Penn94 datasets from the Non-Homophilous Graph Benchmarks suite. All results, besides BOLT, are taken from \cite{lim2021large} and \cite{lim2021new}, except for LinkX on YelpChi, where we used the same experiment setup as those works, including a hyperparameter search. 

From Table \ref{graph-learning}, we see that BOLT achieves state-of-the-art performance on the Yelp-Chi benchmark as well as competitive performance on the other datasets we evaluate. Moreover, our graph learning method trains in one tenth of the time as the baseline algorithms on CPUs. On GPUs, the baseline methods train in roughly the same amount of time as BOLT. Given the significantly larger memory available on modern CPU devices, BOLT provides especially strong value on large-scale graphs that fail to fit in GPU memory.

\begin{table}[h]
\begin{tabular}{|c|c|c|c|}
 \hline  
  & YelpChi & Pokec & Penn94 \\ \hline
MLP & $87.94 \pm 0.52$ & $62.37 \pm 0.02$ & $73.61 \pm 0.40$\\ \hline
GCN & $63.62 \pm 1.00$ & $75.45 \pm 0.17$ & $82.47 \pm 0.27$\\ \hline
GAT & $81.42 \pm 2.12$  & $71.77 \pm 6.18$ & $81.53 \pm 0.55$\\ \hline
LinkX & $77.91 \pm 0.69$ & $82.04 \pm 0.07$ & $84.71 \pm 0.52$ \\ \hline
BOLT & $93.18 \pm 0.45$ & $78.06 \pm 0.07$ & $81.26 \pm 0.40$\\ \hline
\end{tabular}
\captionof{table}{Experimental Results on Large-Scale Non-Homophilous Graph Benchmarks \cite{lim2021large}. YelpChi is evaluated using ROC-AUC while the other datasets use accuracy.}\label{graph-learning}
\vspace{-0.5cm}
\end{table}

\subsection{Carbon Footprint}

As sustainability becomes an increasingly critical requirement for organizations across all business sectors, reducing the cost and energy consumption of training and deploying large-scale neural networks has emerged as a critical task. In the case of GPT-3, for instance, the electricity and compute cost of training alone was reported to be \$12 million~\cite{gpt3_cost}. This concern has only intensified in recent months as model sizes continue to balloon.

To illustrate the energy savings from training with BOLT, we estimate the carbon footprint from running cloud infrastructure using the methodology described in \cite{davy_2021}. For this case study we compare BOLT against RoBERTa~\cite{liu2019roberta}, a state-of-the-art pre-trained transformer model on the Yelp polarity text classification benchmark \cite{zhangCharacterlevelConvolutionalNetworks2015}. We trained BOLT for this task on an AWS r6g.xlarge instance and fine-tuned RoBERTa with a single A100 GPU on a p4dn.24xlarge instance. We provide our carbon footprint estimates using the data in ~\cite{davy_2021} in Table \ref{tab:carbon}. After training, both models achieved the same test accuracy of ~93.3\%. We note that this estimate does not include the pre-training time for RoBERTa, which is a significantly more intensive computational workload than fine-tuning. BOLT, on the other hand, was trained from scratch for this case study with a 10\% level of sparsity. 

\begin{table}[h]
\begin{tabular}{|c|c|c|}
 \hline
 Model & Est. Carbon Footprint & Hourly Cost \\ \hline
 BOLT & 6.1 (g$\text{CO}_2$eq ) & \$0.2240\\ \hline
 RoBERTa & 267.99 (g$\text{CO}_2$eq ) & \$32.773\\ \hline
\end{tabular}
\caption{Estimated carbon footprint of RoBERTA fine-tuning versus BOLT training\label{tab:carbon}}
\vspace{-0.5cm}
\end{table}

\section{Use at Wayfair}

Wayfair is a leading e-commerce company specializing in selling furniture and home goods. With a catalog consisting of tens of millions of products and over thirty million customers, Wayfair relies upon a performant and high quality product search engine to connect a shopper's intent to hyper-relevant products. One component of this search system is a query classifier that maps a search query to the set of products matching the customer's intent, such as dining tables or outdoor chairs. Wayfair previously trained classifiers like this one on GPU hardware before deploying in production on CPUs with a strict latency constraint of a few milliseconds. Motivated by a desire to be able to use larger and more powerful models without compromising on inference latency, the Wayfair data science team was able to train a BOLT model for query classification on low-cost CPU machines and immediately serve the model with no modifications. In online A/B tests, BOLT demonstrated promising results when compared to the baseline production model\footnote{https://www.aboutwayfair.com/careers/tech-blog/how-wayfairs-scientists-collaborated-with-innovative-startup-thirdai-to-serve-hyper-relevant-search-results-to-customers}. 




\section{Conclusion}

We presented BOLT, a production-grade deep learning framework for training and deploying search and recommendation models on commodity CPU hardware. In experimental evaluations, we demonstrate the efficiency and effectiveness of BOLT on a variety of practical machine learning tasks drawn from extreme classification, text classification, personalization, and graph neural networks. We also show
case several key distinguishing features of BOLT, including automated tuning of sparsity hyperparameters and sparse inference. BOLT has also been tested within a leading e-commerce search engine, providing both reduced inference latencies and lower training infrastructure costs.

\bibliographystyle{ACM-Reference-Format}
\bibliography{main}

\end{document}